\documentclass[final]{cvpr}

\usepackage{times}
\usepackage{epsfig}
\usepackage{graphicx}
\usepackage{amsmath}
\usepackage{amssymb}

\usepackage{url}       
\usepackage{booktabs}   
\usepackage{amsfonts}  
\usepackage{nicefrac}
\usepackage{microtype} 
\usepackage{graphicx}
\usepackage{caption}
\usepackage{comment}
\usepackage{multirow}
\usepackage{multicol}
\usepackage{stfloats}

\usepackage[pagebackref=true,breaklinks=true,colorlinks,bookmarks=false]{hyperref}

\begin{document}

\newcommand{\TODO}[1]{{\textcolor{red}{TODO: #1}}}

\title{Towards Part-Based Understanding of RGB-D Scans}

\author{
Alexey Bokhovkin$^{1,2}$\quad\quad~~
Vladislav Ishimtsev$^2$\quad\quad~~
Emil Bogomolov$^2$\quad\quad~~
Denis Zorin$^{3,2}$\\
Alexey Artemov$^2$\quad\quad
Evgeny Burnaev$^2$\quad\quad
Angela Dai$^1$
\vspace{0.3cm} \\ 
$^1$Technical University of Munich~~~\\
$^2$Skolkovo Institute of Science and Technology~~~\\
$^3$New York University
\vspace{0.2cm} \\ 
}

\twocolumn[{%
	\renewcommand\twocolumn[1][]{#1}%
	\maketitle
	\begin{center}
		\vspace{-0.4cm}
		\includegraphics[width=0.95\linewidth]{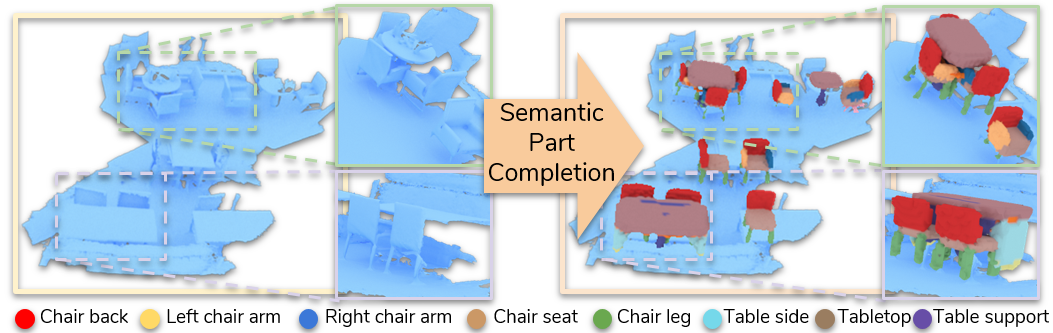}
		\vspace{-0.2cm}
		\captionof{figure}{
		From an input RGB-D scan (left), we propose to detect objects in the scan and predict their complete part decompositions as \emph{semantic part completion}; that is, we predict the part masks for the complete object, inferring the part geometry of any missing or unobserved regions in the scan.
		To achieve this, we predict the part structure of each detected object to drive a geometric prior-driven prediction of the complete part masks.
		}
		\label{fig:teaser}
	\end{center}
}]

\maketitle

\begin{abstract}
   Recent advances in 3D semantic scene understanding have shown impressive progress in 3D instance segmentation, enabling object-level reasoning about 3D scenes; however, a finer-grained understanding is required to enable interactions with objects and their functional understanding.
    Thus, we propose the task of part-based scene understanding of real-world 3D environments: from an RGB-D scan of a scene, we detect objects, and for each object predict its decomposition into geometric part masks, which composed together form the complete geometry of the observed object.
    We leverage an intermediary part graph representation to enable robust completion as well as building of part priors, which we use to construct the final part mask predictions.
    Our experiments demonstrate that guiding part understanding through part graph to part prior-based predictions significantly outperforms alternative approaches to the task of semantic part completion.
\end{abstract}
\section{Introduction}

Recently, we have seen remarkable advances in 3D semantic scene understanding, driven by efforts in large-scale data collection and annotation of 3D reconstructions of RGB-D scanned environments \cite{dai2017scannet,Matterport3D}, coupled with exploration of 3D deep learning approaches across 3D representations such as sparse or dense volumetric grids \cite{wu20153d,qi2016volumetric,dai2017scannet,graham20183dsemantic,choy20194d}, point clouds \cite{qi2017pointnet,qi2017pointnetplusplus}, meshes \cite{gkioxari2019mesh,huang2019texturenet}, and multi-view \cite{dai20183dmv,su2015multi}.
This has led to significant progress in both 3D semantic segmentation as well as 3D semantic instance segmentation \cite{han2020occuseg,graham20183dsemantic,choy20194d,jiang2020pointgroup}.
These have enabled a basis for 3D perception at the level of objects, which is essential for semantic understanding, but lacks finer-grained understanding often critical for enabling interactions with objects and reasoning about functionality (e.g., the seat part of a chair is for sitting on, a knob or handle enables opening doors or drawers).

At the same time, notable progress has been made in part segmentation for shapes \cite{mo2019partnet,mo2019structurenet,hanocka2019meshcnn}.
However, these methods have been developed on synthetic datasets such as ShapeNet~\cite{shapenet2015}, of objects in isolation; this scenario is much less complex than the objects observed in real-world environments.
Thus, we aim to bring these two directions together and propose the task of \emph{semantic part completion}, predicting the part  decomposition of objects in real-world 3D environments, where observations are often cluttered and geometrically incomplete (e.g., due to occlusions, sensor limitations, etc).
That is, from an RGB-D scan of a scene, we detect objects characterized by 3D bounding boxes and class labels, and for each object, we predict its complete part decomposition into binary part masks, with each part mask reflecting the part geometry of the complete object, including unobserved missing regions, to achieve a holistic understanding of the objects in an observed scene.

To achieve this part-based understanding of a scene, we propose to predict the full part graph for each detected object, and based on the predicted part graph, the geometric masks for each complete part.
Predicting the part graph structure enables capturing the complete semantic structure of the object in a low-dimensional representation, allowing reliable prediction of missing and unobserved parts (e.g., for a four-legged table with one leg unobserved, the missing leg is easy to predict based on commonly observed table part patterns).
Furthermore, this enables us to build and exploit strong part geometry priors for each predicted part in the part graph. 
We can then predict the part masks by finding similar part priors and refining them to produce final part mask predictions.
This enables a robust decomposition of an RGB-D scan of a scene into its component objects and their constituent parts, including regions of objects that have been unobserved.
We believe that this takes an important step towards enabling local interactions with objects and functionality analysis in real-world 3D scenes.

We formulate the task of semantic part completion for 3D scene understanding, informing comprehensive part-based object understanding of real-world scans.
To address this part understanding, we propose an approach to decompose a 3D scan of a scene into its complete object parts, outperforming state-of-the-art alternative approaches for the task:
\begin{itemize}
    \item We propose to predict part graph information for objects in real-world scan scenes as an intermediary representation that enables robust, part-based completion of objects.
    \item We leverage the predicted part graphs to guide prior-based prediction for effective inference of geometric part mask decomposition for the objects of a scanned scene.
\end{itemize}

\section{Related Work}

\paragraph{3D Object Detection and Instance Segmentation.}
Following the success of convolutional neural networks for object detection and instance segmentation in 2D images~\cite{girshick2015fast,ren2015faster,redmon2016you,he2017mask}, we are now seeing notable advances in 3D object localization and segmentation.
Earlier approaches leveraging 3D convolutional neural networks developed methods operating on dense voxel grids using 3D region proposal techniques for detection and segmentation~\cite{song2016deep,hou20193dsis}.
Sparse volumetric backbones have also been leveraged to enable effective feature extraction on high-resolution inputs for improved 3D detection and segmentation performance~\cite{engelmann20203d,han2020occuseg}.
Recently, VoteNet~\cite{qi2019deep} introduced a Hough Voting-inspired scheme for 3D object detection on point clouds.
This was extended by MLCVNet~\cite{xie2020mlcvnet} to incorporate multi-scale contextual information for improved detection performance.
These approaches have now shown impressive performance for instance-level scene understanding; we aim to build upon this and propose to infer finer-grained part decomposition for each object in a 3D scan.

\paragraph{3D Scan Completion.}
Repairing and completing holes or broken meshes has been well-studied for 3D shapes.
Traditional methods have mainly focused on repairing small holes by fitting geometric primitives, continuous energy minimization, or leveraging surface reconstruction for interpolation of missing regions~\cite{nealen2006laplacian,zhao2007robust,speciale2016symmetry,kazhdan2006poisson,kazhdan2013screened}.
Structural or symmetry priors have also been leveraged for shape completion~\cite{thrun2005shape,mitra2006partial,pauly2008discovering,sipiran2014approximate,speciale2016symmetry}.
Recently, generative deep learning approaches have been developed, with significant progress in 3D shape reconstruction and completion~\cite{wu20153d,dai2017shape,hane2017hierarchical,park2019deepsdf}.

In addition to operating on the limited spatial context of shapes, generative deep learning approaches have also been developed for completion of 3D scenes.
Song et al.~\cite{song2017semantic} developed a voxel-based approach to predict  geometric occupancy of a single depth frame, leveraging a large-scale synthetic 3D dataset of scenes.
Dai et al.~\cite{dai2018scancomplete} proposed an autoregressive approach for scan completion, enabling very large scale completion.
SG-NN~\cite{dai2020sgnn} presented a self-supervised approach towards 3D scan completion, enabling training only on real scan data.
These approaches operate on geometric completion but without knowledge of individual object instances, which is fundamental to many perception-based tasks.
RevealNet~\cite{hou2020revealnet} introduced an approach to detect objects in a 3D scan and infer each object's complete geometry, joining together geometric reconstruction with object-based understanding.
We similarly aim to infer each object's complete geometry from a partial scan observation, but infer a part decomposition of the object structure, enabling both finer-grained understanding as well as more effective object completion through its part structure.

\paragraph{Part Segmentation of 3D Shapes.}
Understanding the structure of a 3D shape by identifying shape parts has been long-studied in shape analysis.
Various approaches have been developed for finding a consistent segmentation across a set of shapes without supervision of part labels \cite{golovinskiy2009consistent,huang2011joint,sidi2011unsupervised,hu2012co}.
Recently, deep learning based approaches have been developed to find part segmentation of shapes in a data-driven fashion~\cite{kalogerakis20173d,yi2016scalable,hanocka2019meshcnn}.
To better capture more complex structures in the part layout of shapes, several methods propose to parse object parts as hierarchies \cite{wang2011symmetry,van2013co,yi2017learning,mo2019partnet,mo2019structurenet}.
Such hierarchically structured representations have also been adopted for 3D scene synthesis, leveraging a scene graph  \cite{fisher2012example,zhao2016relationship, li2019grains}, where object instances rather than parts form the node primitives.
We also adopt a hierarchically structured approach towards part decomposition, but aim to operate on noisy, incomplete real-world scans of scenes with multiple objects, and so propose to combine our hierarchical part decomposition with strong geometric part priors.

\section{Method}

\begin{figure*}
\begin{center}
	\includegraphics[width=\linewidth]{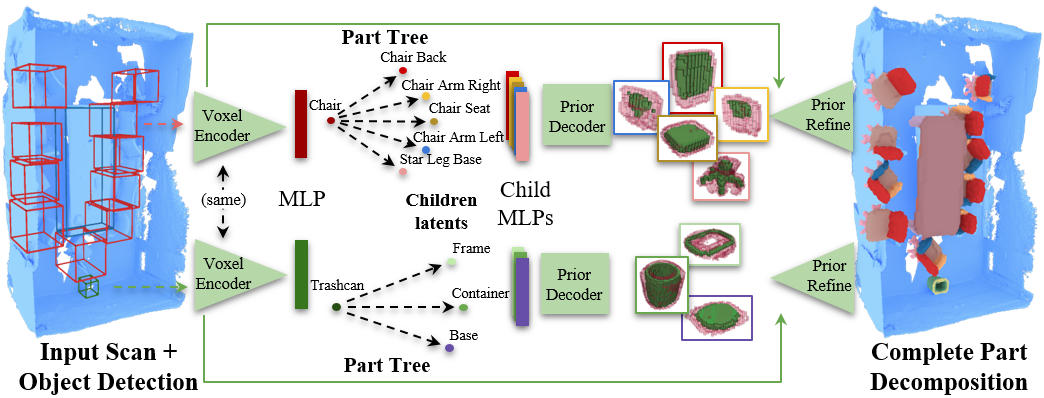}
	\vspace{-0.5cm}
   \caption{Overview of our approach. From an input scan, we detect objects as their 3D bounding boxes, and for each object (a chair and a trash can visualized top and bottom, respectively), we predict their part tree structure, which is then used to guide a geometric prior-based part mask prediction. This results in a part decomposition of the scene where each object is decomposed into its complete part geometry, including any missing or unobserved regions.}
	\vspace{-0.5cm}
\label{fig:overview}
\end{center}
\end{figure*}

\subsection{Overview}

We address the problem of simultaneous part segmentation and completion of objects of real-world RGB-D scans, which are often noisy and incomplete.
An overview of our approach is illustrated on Fig.~\ref{fig:overview}. 
Given an input 3D scan $\mathbb{S}$, we aim to predict a set of parts for each object in the scan, with each part representing the complete geometry of the part, including any missing or unobserved regions.
From $\mathbb{S}$, we first detect a set of object instances $\mathbb{O}$ = \{$o_i$\} in the scene,  as 3D bounding box locations and class category predictions. 
For each detected object in $\mathbb{O}$, we then convert it into a $32^3$ occupancy grid representation, to inform our part segmentation and completion. 

We then predict the part segmentation and completion for each detected object $o_i\in \mathbb{O}$.
First, for a detected object $o_i$, we predict its full part decomposition as a tree graph $T_i$ corresponding to the part hierarchy structure (nodes representing the part class types), with part hierarchies derived from those of PartNet~\cite{mo2019partnet}.
This enables encoding the high-level, semantic part structure of the shape, which both facilitates completion of the shape structure, as missing parts are easy to identify in their part tree structure, as well as guides the prediction of the geometry of each part.
In particular, this allows us to leverage geometric part priors built for each part category.
We construct the part priors based on clustering of train part masks for each part category, and learn to predict similar priors for each leaf in our predicted $T_i$, followed by a refinement of these priors to predict the final part mask geometry.
This produces a semantic part decomposition of objects in a 3D scan while simultaneously inferring their complete part geometry.

\subsection{Object Detection}
From an input 3D scan, we first detect objects in the scene.
We leverage a state-of-the-art 3D object detection approach, MLCVNet~\cite{xie2020mlcvnet}, as our object detection backbone.
The input scan sampled to a point cloud, and object proposals are produced by voting~\cite{qi2019deep}, leveraging global contextual information at various scales.
As output, we obtain 3D bounding box locations for each detected object.
We then resample the input scan geometry within each detected box into $32^3$ occupancy grids $o_i\in \mathbb{O}$ to inform our part decomposition.

For a detected object $o_i$ from the scan, represented as a $32^3$ occupancy grid of the scan geometry within its predicted bounding box, we encode the occupancy grid with  four 3D convolutional blocks (consisting of convolution, group normalization and ReLU activation) and extract a feature encoding $z_i$ of dimension $128$, which is used to inform the part decomposition. 

\paragraph{Object Orientation Prediction}
Since our object detection backbone predicts axis-aligned bounding boxes for each object, we additionally predict the orientation $r_i$ of each object $o_i$ from its feature $z_i$ using an MLP.
We assume that the up (gravity) vector is known in the scene, and thus predict the angle around the up vector by classifying the angle in $n_{\alpha}=8$ bins of discretized angles ($\{0^{\circ}, 45^{\circ}, \dots, 315^{\circ}\}$) with a cross entropy loss.
The predicted object orientation helps to guide our prior-based part decomposition as described in Section~\ref{subsec:partmask}.

\subsection{Part Tree Prediction}

For a detected object $o_i$ from the scan, represented as a $32^3$ occupancy grid of the scan geometry within its predicted bounding box, we aim to capture its high-level part structure from its cluttered and partial observation.
We predict the part tree structure of the object; this facilitates completion of the object by predicting its high-level structure, as well as enables our prior-guided part geometry prediction.

We first encode the occupancy grid of $o_i$ with four 3D convolutional blocks (consisting of convolution, group normalization and ReLU activation), and extract a feature encoding $z_i$ of dimension $128$.
We then decode $z_i$ into a part tree prediction, constructing a part tree $T_i$ with each node represented by its predicted part category and a 128-dimensional feature encoding.
Inspired by StructureNet~\cite{mo2019structurenet}, we leverage a message-passing graph neural network for our part tree prediction.
From $z_i$, we predict tree children nodes using an MLP to predict $n_{children} = 10$ latent vectors $\{z'_k\}$ that correspond to potential parts of $o$.
We additionally predict a tuple $t_k = (e_k, s_k)$ for every child $z'_k$, where $e_k$ is the probability of child existence, $s_k$ is the one-hot representation of the part category label. 
For each pair $(z'_i, z'_j)$ of nodes, we predict if they are adjacent or not, enforcing structural features to be learned by the message-passing network.
We employ a cross entropy loss for the part category label, and binary cross entropy losses for node existence and adjacency relationships. 
This produces a high-level part summary of $o_i$, where nodes $\{z'_k\}$ represent part semantic information of the complete structure of $o_i$, even if $o_i$ has been partially observed.
We leverage this part semantic information to guide our final part decomposition as geometric part masks.

\subsection{Prior-guided Part Decomposition}\label{subsec:partmask}

\begin{figure}
\begin{center}
	\includegraphics[width=0.95\linewidth]{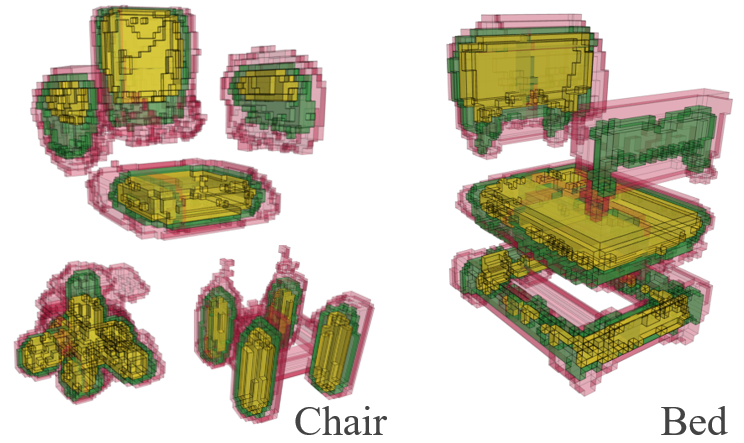}
	\vspace{-0.3cm}
   \caption{Several geometric part priors for part types belonging to the `chair' and `bed' class categories. Each part prior represents a cluster of train parts, visualized at three different isolevels.}
\label{fig:partpriors}
\end{center}
\end{figure}

We then predict the final part decomposition by generating part masks for each node in the predicted part tree $T_i$, where each mask represents the complete geometry associated with the part, including regions that were unobserved in the initial scan observation.
Rather than directly reconstructing the part geometry of each predicted part in the part tree, we observe that object parts often maintain very similar geometry structures, which we leverage to obtain our final part decomposition.
That is, we construct geometric part priors to aid in generating our complete part mask predictions, and learn to find similar geometric part priors which we then refine for a final prediction.

We construct our geometric part priors by $k$-means clustering of the binary part masks in the train set, inspired by the ShapeMask~\cite{kuo2019shapemask} approach to building priors for novel 2D object segmentation.
For each part type, we find $K=10$ centroids of the part masks of that type, and perform the clustering on the part masks in $32^3$ grids of the canonical object space.
This produces a set of part priors $\{P_1, \dots, P_{M}\}$ with $M = n_{\textrm{classes}}K$.
Various resulting part priors are visualized in Figure~\ref{fig:partpriors}.
Since objects in the real-world scan inputs may not be oriented in the canonical orientation of the object, we use the predicted orientation $r_i$ to transform the priors to $\{P_1^r, \dots, P_{M}^r\}$.

Thus, to predict the part geometry associated for a node in the predicted part tree $T_i$ with feature encoding $z'_k$ and predicted part type $t$, we use a one-layer MLP which takes as input $z'_k$ and predicts a set of weights $w_m$ used to construct an initial part reconstruction as:
$$P^{\textrm{coarse}}_k = \sum_{m=1}^{M_t} w_mP_m^r,$$
where $w = softmax(\phi(z'_k))$, and $\phi$ is a linear layer.
We employ a proxy loss on this initial part reconstruction, using a mean squared error with a target part mask.

Such prior-guided part decomposition helps to reconstruct global structures in part masks such as symmetry and geometry in missing regions in the input observation.
We then refine the predicted $P^{\textrm{coarse}}_k$ using four 3D convolutional blocks (consisting of convolution, batch normalization and ReLU activation) taking as input the concatenation of the geometry of $o_i$ and $P^{\textrm{coarse}}_k$ to produce $P^{\textrm{refine}}_k$; we then obtain the final part mask prediction, $$P_k = P^{\textrm{coarse}}_k + P^{\textrm{refine}}_k.$$
Empirically, we found that predicting the refinement as residuals to modify the initial $P^{\textrm{coarse}}_k$ to perform better than a direct refinement (c.f. Section~\ref{sec:results}).
We then employ a binary cross entropy loss on $P_k$ with a target part mask.
This encourages an improved local fit to the observed geometry that may not have been captured in the global structure of the geometric priors.

\subsection{Training Details}

\paragraph{Data generation.}
In order to train our approach, we leverage the Scan2CAD dataset~\cite{avetisyan2019scan2cad} in combination with PartNet~\cite{mo2019partnet}.
Scan2CAD contains annotations of CAD models from ShapeNet~\cite{shapenet2015} aligned to the 3D scans of ScanNet~\cite{dai2017scannet}, and we use the part annotations of PartNet for these ShapeNet CAD models to obtain our ground truth part decompositions of the 3D scans.
We leverage the ground truth CAD alignments to compute our geometric part priors in the canonically-oriented space of the objects, and use our rotation prediction during training and inference to orient them to the scan observations. In all our experiments we use original ScanNet geometry with typical number of points as 200k per scene, for MLCVNet method 40k points are randomly sampled from each scene to train object detection.

\paragraph{Training.}
We train our part tree prediction and geometric decomposition model with an Adam optimizer, using a batch size of $24$, learning rate of $0.001$, and weight decay of $0.01$. 
The learning rate is decayed every $8$ epochs by a factor of $0.8$.
We first pre-train for $20$ epochs using ground truth 3D bounding boxes, and then fine-tune for $10$ epochs with geometry from MLCVNet detections.
MLCVNet is trained using the original proposed parameters: using an Adam optimizer with batch size $8$, learning rate $0.01$, for $250$ epochs.

\begin{table*}[tp]
\centering
\resizebox{\textwidth}{!}{
\begin{tabular}{l|cccccc|cc||cccccc|cc}
& \multicolumn{8}{c||}{Chamfer Distance ($\downarrow$)} & \multicolumn{8}{c}{IoU ($\uparrow$)}\\
\toprule
    Method & chair & table & cab. & bkshlf & bed & bin & class avg & inst avg & chair & table & cab. & bkshlf & bed & bin & class avg & inst avg \\
\midrule
    SG-NN + MLCVNet + UNet & 0.050 & 0.118 & 0.080 & 0.053 & 0.083 & 0.108 & 0.082 & 0.073 & 17.5 & 6.4 & 7.6 & 12.4 & 13.3 & 13.9 & 11.9 & 13.3 \\
    SG-NN + MLCVNet + PointGroup & 0.074 & 0.102 & 0.100 & 0.063 & 0.091 & 0.140 & 0.095 & 0.093 & 5.1 & 1.5 & 1.0 & 4.5 & 4.5 & 0.9 & 2.9 & 2.9 \\
    MLCVNet + StructureNet & {\bf 0.029} & 0.095 & {\bf 0.065} & 0.037 & 0.076 & 0.106 & 0.068 & 0.057 & 13.8 & 0.5 & 3.8 & 9.0 & 3.9 & 9.3 & 6.8 & 8.9 \\
\midrule
    {\bf Ours} & 0.033 & {\bf 0.089} & 0.069 & {\bf 0.033} & {\bf 0.054} & {\bf 0.096} & {\bf 0.062} & {\bf 0.053} & {\bf 22.1} & {\bf 7.7} & {\bf 13.0} & {\bf 18.1} & {\bf 17.3} & {\bf 22.0} & {\bf 16.7} & {\bf 18.3} \\
\bottomrule
\end{tabular}
}
\vspace{-0.2cm}
\caption{Evaluation on semantic part completion on Scan2CAD~\cite{avetisyan2019scan2cad}. We compare with state-of-the-art approaches for scan completion~\cite{dai2020sgnn}, followed by object detection~\cite{xie2020mlcvnet}, and then part segmentation \cite{jiang2020pointgroup, mo2019structurenet}.
By leveraging part structures to guide our prior-based approach, we obtain more accurate part decompositions.
}
\label{tab:part_cmpl_comparison}
\end{table*}

\begin{table*}
\centering
\resizebox{\textwidth}{!}{
\begin{tabular}{l|cccccc|cc||cccccc|cc}
& \multicolumn{8}{c||}{Chamfer Distance ($\downarrow$)} & \multicolumn{8}{c}{IoU ($\uparrow$)}\\
\toprule
    Method & chair & table & cab. & bkshlf & bed & bin & class avg & inst avg & chair & table & cab. & bkshlf & bed & bin & class avg & inst avg \\
\midrule
    MLCVNet + UNet & 0.052 & 0.082 & {\bf 0.062} & 0.034 & 0.093 & 0.068 & 0.065 & 0.060 & 24.1 & 13.4 & 9.3 & {\bf 31.8} & 14.3 & 14.6 & 17.9 & 18.9 \\
    MLCVNet + PointGroup & 0.054 & {\bf 0.057} & 0.077 & 0.045 & {\bf 0.072} & 0.086 & 0.065 & 0.061 & 28.4 & 14.9 & 9.6 & 27.5 & 18.8 & 11.9 & 18.5 & 19.6 \\
    MLCVNet + StructureNet & {\bf 0.039} & 0.084 & {\bf 0.062} & 0.034 & 0.075 & 0.083 & 0.063 & 0.056 & {\bf 32.6} & 2.1 & 9.4 & 23.1 & 16.1 & 15.4 & 16.5 & 15.4 \\
\midrule
    {\bf Ours} & 0.044 & 0.072 & 0.063 & {\bf 0.031} & 0.092 & {\bf 0.063} & {\bf 0.061} & {\bf 0.054} & 30.9 & {\bf 16.5} & {\bf 10.9} & {\bf 31.8} & {\bf 20.6} & {\bf 20.9} & {\bf 21.9} & {\bf 24.7} \\
\bottomrule
\end{tabular}
}
\vspace{-0.3cm}
\caption{Evaluation of part segmentation on Scan2CAD~\cite{avetisyan2019scan2cad}. We evaluate part segmentation of visible geometry only, in comparison with state-of-the-art part segmentation \cite{jiang2020pointgroup, mo2019structurenet}. }
\label{tab:part_seg_comparison}
\end{table*}

\section{Results}
\label{sec:results}

\begin{figure*}
\begin{center}
	\includegraphics[width=0.88\linewidth]{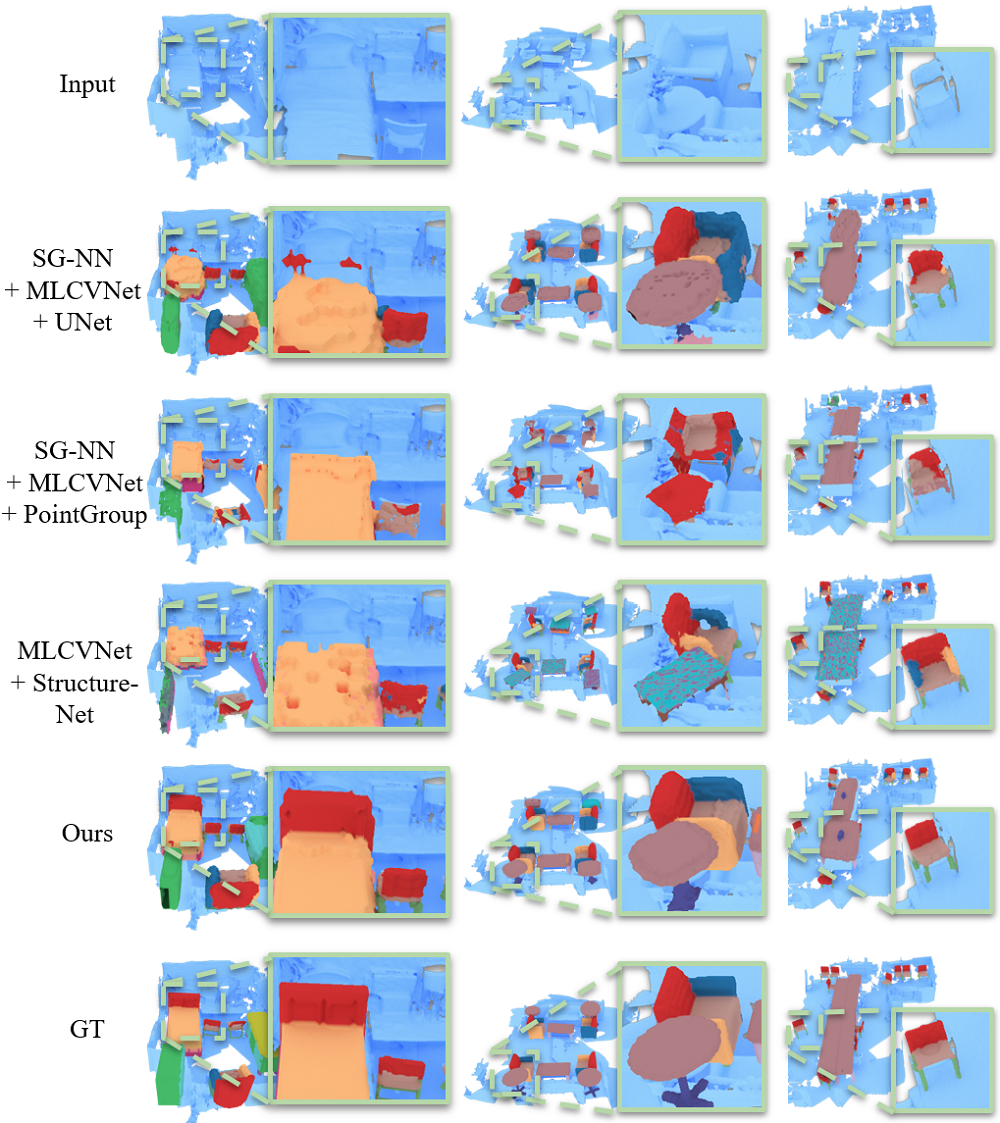}
	\vspace{-0.2cm}
   \caption{Qualitative evaluation on semantic part completion in comparison with state-of-the-art approaches for part decomposition, including scan completion followed by part segmentation. Our approach produces more consistent, accurate part decompositions.}
\label{fig:part_cmpl_comparison}
\end{center}
\end{figure*}

We evaluate our proposed approach in comparison to alternative approaches for semantic part completion on real-world RGB-D scans.
We use scans from the ScanNet dataset~\cite{dai2017scannet}, containing 1513 reconstructed RGB-D scans, and evaluate with their train/val/test split of 1045/156/312 scenes, respectively.
To train and evaluate the complete part decomposition for each object, we use the Scan2CAD~\cite{avetisyan2019scan2cad} annotations of CAD model alignments from ShapeNet~\cite{shapenet2015} to the ScanNet scans, coupled with the PartNet~\cite{mo2019partnet} annotations for the part decomposition of the ShapeNet CAD models.
We train and evaluate on 6 object class categories representing the majority of parts (45 part types in total that we train and evaluate on) for these annotations.
For a detailed specification of the part types used, we refer to the appendix.

To evaluate our part decompositions of the objects in a scan, we use a Chamfer Distance metric as well as an intersection over union (IoU) metric.
For IoU, we evaluate $32^3$ voxelizations of each predicted part in object space, compared to the Scan2CAD ground truth part.
For Chamfer Distance, we use the predicted voxel centers as points, normalized to the unit box of the object. 
For both Chamfer Distance and IoU, we compute the metrics for each part type and average over all part types corresponding to an object class category. 
The class average is computed by averaging all resulting category numbers, and instance average computed by averaging the metrics of all part instances regardless of their object category. 
Note that to evaluate part segmentation without completion, we consider only predictions which overlap with the original scan geometry.

\begin{table*}[tp]
\centering
\resizebox{\textwidth}{!}{
\begin{tabular}{l|cccccc|c||cccccc|c}
& \multicolumn{7}{c||}{Chamfer Distance ($\downarrow$)} & \multicolumn{7}{c}{IoU ($\uparrow$)}\\
\toprule
    Method & chair & table & cab. & bkshlf & bed & bin & {\bf avg} & chair & table & cab. & bkshlf & bed & bin & {\bf avg} \\
\midrule
    MLCVNet + StructureNet & {\bf 0.0032} & 0.0106 & 0.0074 & 0.0046 & 0.0194 & {\bf 0.0025} & 0.0079 & 29.5 & 21.9 & 22.4 & 24.1 & 23.5 & 32.4 & 25.6\\
    RevealNet & 0.0035 & {\bf 0.0070} & 0.0043 & {\bf 0.0020} & 0.0076 & 0.0078 & 0.0053  & 35.1 & 26.2 & 46.1 & {\bf 38.3} & 19.6 & 24.7 & 31.7\\
    MLCVNet + UNet & 0.0038 & 0.0103 & {\bf 0.0011} & 0.0050 & 0.0119 & 0.0028 & 0.0059 & {\bf 39.7} & 30.0 & {\bf 62.6} & 28.2 & 17.4 & 37.7 & 35.9\\
\midrule
    {\bf Ours} & 0.0038 & 0.0075 & 0.0022 & 0.0053 & {\bf 0.0061} & 0.0045 &  {\bf 0.0049} & 38.6 & {\bf 30.7} & 57.4 & 33.6 & {\bf 37.6} & {\bf 37.8} & {\bf 39.3}\\
\bottomrule
\end{tabular}
}
\vspace{-0.2cm}
\caption{Evaluation of instance completion on Scan2CAD~\cite{avetisyan2019scan2cad}. We evaluate object completion as a union of predicted part decompositions, in comparison with state-of-the-art instance completion~\cite{hou2020revealnet} and the union of StructureNet~\cite{mo2019structurenet} parts as instances.}
\label{tab:inst_cmpl_comparison}
\end{table*}

\paragraph{Comparison to alternative approaches.}
In Table~\ref{tab:part_cmpl_comparison}, we compare to several state-of-the-art approaches for part segmentation and scan completion, coupled together to provide a complete part decomposition of the objects in a scan.
As an alternative approach for this task, we consider scan completion followed by object detection and part instance segmentation.
We employ the state-of-the-art scan completion approach SG-NN~\cite{dai2020sgnn} to generate a prediction for the complete geometry of a partial scan observation, then detect object instances with MLCVNet~\cite{xie2020mlcvnet}, obtain a final  complete part decomposition  by the state-of-the-art instance segmentation of PointGroup~\cite{jiang2020pointgroup}.
We also compare to StructureNet~\cite{mo2019structurenet} on MLCVNet detections, following their approach of using a pretraining a decoder for complete part decompositions and then learning an encoder to map this space. 
We additionally consider a UNet~\cite{ronneberger2015u} composed of 3D volumetric convolutions as a baseline for the final part segmentation.
We train these alternative approaches on our part decomposition data for ScanNet.
These approaches do not consider explicit part structure reasoning, whereas our part tree prediction guiding our part decomposition with geometric priors enables a more effective complete part decomposition.

In Figure~\ref{fig:part_cmpl_comparison}, we show a qualitative comparison: without part structure reasoning, the PointGroup approach can often mix up geometrically similar parts such as the left and right chair arms, and the UNet baseline suffers in generating complete part structures. 
StructureNet provides part structure reasoning, but their approach to train an encoder into a pretrained decoder can tend to predict only the dominant part decompositions for a class category (e.g., an office-type chair instead of an armchair in the third row of Figure~\ref{fig:part_cmpl_comparison}). 
Our part structure guided priors enable more effective and accurate part decompositions of the objects in the scenes.

\begin{table*}[bp]
\centering
\resizebox{\textwidth}{!}{
\begin{tabular}{l|cccccc|cc||cccccc|cc}
& \multicolumn{8}{c||}{Chamfer Distance ($\downarrow$)} & \multicolumn{8}{c}{IoU ($\uparrow$)}\\
\toprule
    Method & chair & table & cab. & bkshlf & bed & bin & class avg & inst avg & chair & table & cab. & bkshlf & bed & bin & class avg & inst avg \\
\midrule
    w/o Part Msg Pass & 0.037 & 0.094 & 0.069 & 0.039 & 0.077 & 0.096 & 0.069 & 0.057 & 20.0 & 6.6 & 9.8 & 14.2 & 14.4 & 20.6 & 14.3 & 16.3 \\
    w/o Priors & 0.036 & 0.093 & 0.067 & 0.044 & 0.058 & 0.101 & 0.067 & 0.056 & 21.8 & 7.3 & 11.0 & 13.8 & 16.4 & 21.9 & 15.4 & 17.7 \\
    No Prior Refine & 0.034 & 0.093 & 0.069 & 0.034 & 0.057 & 0.096 & 0.064 & 0.055 & {\bf 22.5} & 7.6 & 12.2 & 17.9 & 16.6 & {\bf 22.0} & 16.4 & 18.2 \\
    Prior Refine (Abs) & 0.036 & {\bf 0.089} & {\bf 0.065} & 0.034 & 0.067 & 0.105 & 0.066 & 0.055 & 21.4 & 7.5 & 11.5 & 17.4 & 16.5 & 20.7 & 15.8 & 17.6 \\
    {\bf Ours} & {\bf 0.033} & {\bf 0.089} & 0.069 & {\bf 0.033} & {\bf 0.054} & {\bf 0.096} & {\bf 0.062} & {\bf 0.053} & 22.1 & {\bf 7.7} & {\bf 13.0} & {\bf 18.1} & {\bf 17.3} & {\bf 22.0} & {\bf 16.7} & {\bf 18.3} \\
\bottomrule
\end{tabular}
}
\vspace{-0.2cm}
\caption{Ablation study for our design decisions, evaluated for semantic part completion on Scan2CAD~\cite{avetisyan2019scan2cad}.}
\label{tab:ablations}
\end{table*}

\paragraph{Part segmentation on 3D scans.}
In addition to our task of semantic part completion, we evaluate our approach in comparison to state of the art on part segmentation  in Table~\ref{tab:part_seg_comparison}.
To evaluate part segmentation, we consider only the part predictions that intersect with the original scan geometry, and compare to PointGroup~\cite{jiang2020pointgroup}, StructureNet~\cite{mo2019structurenet}, and a UNet baseline, using the object detection of by MLCVNet~\cite{xie2020mlcvnet}.
For part segmentation, we see that our part structure reasoning coupled with geometric priors also produces more consistent part segmentations of the objects in a scan.

\begin{figure*}[tp]
\begin{center}
	\includegraphics[width=0.9\linewidth]{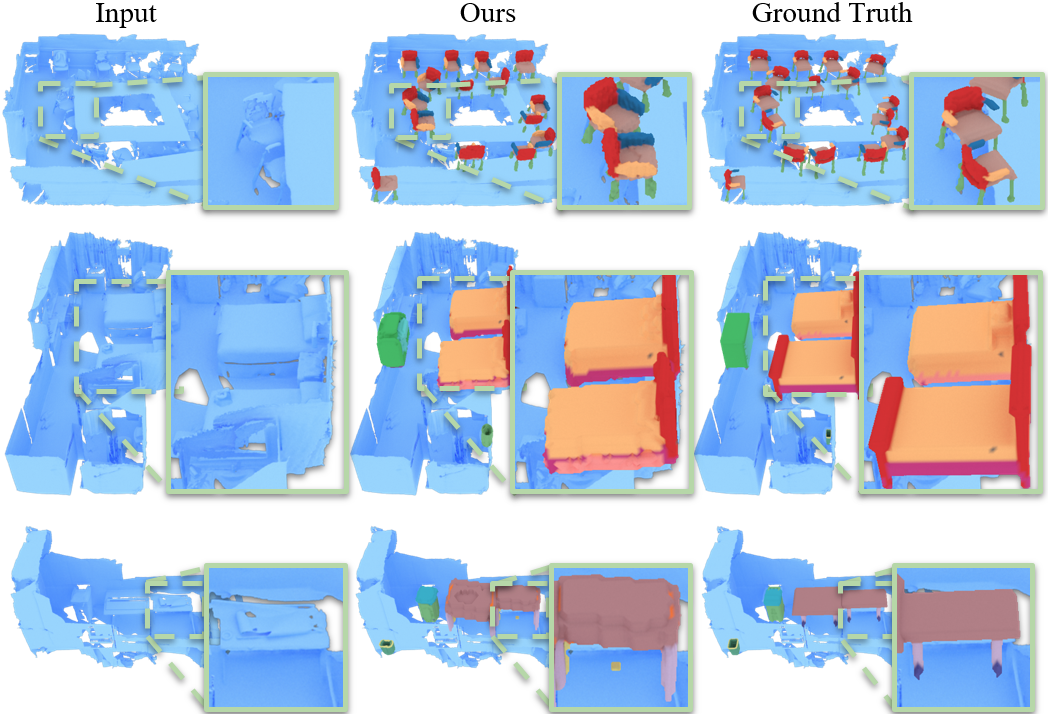}
	\vspace{-0.1cm}
   \caption{Qualitative results on real-world ScanNet~\cite{dai2017scannet} scenes using Scan2CAD~\cite{avetisyan2019scan2cad} and PartNet~\cite{mo2019partnet} targets. Our approach effectively predicts each object's complete geometry as a decomposition into semantic parts.}
\label{fig:gallery}
\end{center}
\end{figure*}

\paragraph{Object completion on 3D scans.} 
In Table~\ref{tab:inst_cmpl_comparison}, we additionally evaluate our approach on object instance completion by taking the union of our part mask predictions as a complete object mask prediction. 
We compare to RevealNet~\cite{hou2020revealnet}, which established this task, as well as a state-of-the-art object detection using MLCVNet~\cite{xie2020mlcvnet} followed by a UNet for completion or by StructureNet~\cite{mo2019structurenet}.
Our part reasoning enables more effective instance completion by explicitly leveraging shared structural knowledge of objects.

\paragraph{Ablations.}
In Table~\ref{tab:ablations}, we analyze the effect of our design decisions for part tree and prior-guided part mask prediction.
We evaluate our approach without message-passing in our part tree prediction (\emph{w/o Part Msg Pass}), without using priors and directly decoding with convolutions to a part mask prediction (\emph{w/o Priors}), without refinement of priors (\emph{No Prior Refine}), and prior refinement with absolute predictions instead of our relative offsets that are added to the raw prior prediction (\emph{Prior Refine (Abs)}).
Our prior-guided predictions, with refinement learned as a residual offset, helps to produce more accurate results.

We additionally consider the effect of varying voxel resolutions in Table~\ref{tab:varying_resolutions}. 
All resolutions produce meaningful results, although a (twice) higher resolution can result in somewhat noisier results, and a (half) lower resolution tends to lack detail. 
Thus we choose to employ a $32^3$ resolution for each object.

\paragraph{Limitations.}
While our approach for semantic part completion shows promise towards a finer-grained, semantically part-based understanding of 3D environments, we believe there are many avenues for further development. 
For instance, a dense volumetric representation of parts may suffice for functionality analysis of furniture-type objects, but can struggle to generate very high resolution parts for small objects; we believe sparse~\cite{graham20183dsemantic,choy20194d} or hierarchical~\cite{riegler2017octnet,tatarchenko2017octree} approaches would complement our prior-based approach.
Furthermore, objects are currently considered independently for each part decomposition, where relational inference between objects in a scene would help to explain noisy or unobserved part regions (e.g., multiple chairs or tables in a scene are often repeated instances of the same geometry).

\begin{table*}[bp]
\centering
\resizebox{\textwidth}{!}{
\begin{tabular}{l|cccccc|cc||cccccc|cc}
& \multicolumn{8}{c||}{Chamfer Distance ($\downarrow$)} & \multicolumn{8}{c}{IoU ($\uparrow$)}\\
\toprule
    Method & chair & table & cab. & bkshlf & bed & bin & class avg & inst avg & chair & table & cab. & bkshlf & bed & bin & class avg & inst avg \\
\midrule
    Res. 16  & 0.034 & 0.088 & 0.072 & 0.054 & 0.061 & 0.109 & 0.070 & 0.055 & 28.4 & 10.5 & 13.5 & 20.9 & 18.5 & 21.2 & {\bf 18.8} & {\bf 22.8} \\
    Res. 32  & 0.033 & 0.089 & 0.069 & 0.033 & 0.054 & 0.096 & {\bf 0.062} & {\bf 0.053} & 22.1 & 7.7 & 13.0 & 18.1 & 17.3 & 22.0 & 16.7 & 18.3 \\
    Res. 64  & 0.045 & 0.098 & 0.058 & 0.044 & 0.067 & 0.100 & 0.069 & 0.060 & 18.8 & 5.6 & 9.9 & 10.5 & 14.7 & 19.3 & 13.1 & 15.4 \\
\bottomrule
\end{tabular}
}
\vspace{-0.2cm}
\caption{Evaluation of various object resolutions during training for semantic part completion on Scan2CAD~\cite{avetisyan2019scan2cad}.}
\label{tab:varying_resolutions}
\end{table*}
\section{Conclusion}

In this paper, we have presented a new approach for the semantic part completion task of predicting a geometrically complete part decomposition for each object in a 3D scan.
For each detected object in a scene, we exploit explicit part structure prediction in order to guide a geometric part prior prediction, which is then refined to a final part decomposition, where each part is represented by its semantic part type as well as the geometry corresponding to the part, including any missing or unobserved regions in the scan.
We show that our structural and prior-guided reasoning about object parts notably outperforms alternative approaches on this task.
We believe that our approach makes an important step towards part-based understanding of 3D environments, and opens up new possibilities for part-level functionality analysis, autonomous agent interactions with an environment, and more.
\section{Acknowledgements}

We would like to thank the support of the Zentrum Digitalisierung.Bayern (ZD.B) and the Russian Science Foundation (Grant 19-41-04109). We additionally acknowledge the use of Skoltech CDISE HPC cluster Zhores.

{\small
\bibliographystyle{ieee_fullname}
\bibliography{egbib}
}

\clearpage

\appendix
\section{Appendix}

In this appendix,
we detail our network architecture in Section~\ref{sec:app_network}; in Section~\ref{sec:app_baselines}, we provide details of our baselines designs; in Section~\ref{sec:app_parts}, we provide specifications of parts that we used in our experiments; in Section~\ref{sec:app_results}, we additionally provide more quantitative results, visualize examples of part priors combinations for each main category and examples of our predictions compared to ground-truth.

\begin{table}[bp]
\centering
\resizebox{0.5\textwidth}{!}{
\begin{tabular}{l|cccccc|c}
& \multicolumn{7}{c}{mAP@25 ($\uparrow$)} \\
\toprule
    Method & chair & table & cab. & bkshlf & bed & bin & {\bf avg} \\
\midrule
    MLCVNet + StructureNet & 45.7 & 25.7 & 19.8 & 50.0 & 36.4 & 53.0 & 38.4 \\
    RevealNet & 70.3 & 40.6 & 90.5 & 87.2 & 22.7 & 20.6 & 55.3 \\
\midrule
    Ours & 78.4 & 47.2 & 90.5 & 77.8 & 22.7 & 72.4 & 64.8 \\
\bottomrule
\end{tabular}
}
\vspace{-0.2cm}
\caption{Evaluation of instance completion on Scan2CAD~\cite{avetisyan2019scan2cad}. We evaluate object completion as a union of predicted part decompositions, in comparison with state-of-the-art instance completion~\cite{hou2020revealnet} and the union of StructureNet~\cite{mo2019structurenet} parts as instances.}
\label{tab:inst_cmpl_map_comparison}
\end{table}

\begin{table*}[bp]
\centering
\resizebox{\textwidth}{!}{
\begin{tabular}{l|cccccc|cc||cccccc|cc}
& \multicolumn{8}{c||}{Chamfer Distance ($\downarrow$)} & \multicolumn{8}{c}{IoU ($\uparrow$)}\\
\toprule
    Method & chair & table & cab. & bkshlf & bed & bin & class avg & inst avg & chair & table & cab. & bkshlf & bed & bin & class avg & inst avg \\
\midrule
    StructureNet~\cite{mo2019structurenet} & 0.019 & 0.089 & 0.048 & 0.032 & 0.069 & 0.105 & 0.061 & 0.049 & 18.5 & 1.0 & 10.1 & 16.8 & 6.8 & 12.1 & 10.9 & 12.8 \\
    {\bf Ours} & 0.029 & 0.089 & 0.055 & 0.037 & 0.058 & 0.081 & 0.058 & 0.048 & 27.6 & 8.0 & 17.3 & 20.9 & 19.8 & 28.7 & 20.4 & 22.6 \\
\bottomrule
\end{tabular}
\vspace{-0.4cm}
}
\caption{Evaluation on semantic part completion on Scan2CAD~\cite{avetisyan2019scan2cad} with ground truth 3D object detection (oriented 3D bounding boxes) as input.}
\label{tab:part_cmpl_comparison_gt}
\end{table*}

\begin{table*}[tp]
\centering
\resizebox{0.8\textwidth}{!}{
\begin{tabular}{c|c|c|c|c|c|c|c}
\toprule
    Encoder & Input Layer & Type & Input Size & Output Size & Kernel Size & Stride & Padding \\
\midrule
    conv0 & scan occ. grid & Conv3D & (1, 32, 32, 32) & (16, 16, 16, 16) & (5, 5, 5) & (2, 2, 2) & (2, 2, 2) \\
    gnorm0 & conv0 & GroupNorm & (16, 16, 16, 16) & (16, 16, 16, 16) & - & - & - \\
    relu0 & gnorm0 & ReLU & (16, 16, 16, 16) & (16, 16, 16, 16) & - & - & - \\
    pool1 & relu0 & MaxPooling & (16, 16, 16, 16) & (16, 8, 8, 8) & (2, 2, 2) & (2, 2, 2) & (0, 0, 0) \\
    conv1 & pool1 & Conv3D & (16, 8, 8, 8) & (32, 8, 8, 8) & (3, 3, 3) & (1, 1, 1) & (1, 1, 1) \\
    gnorm1 & conv1 & GroupNorm & (32, 8, 8, 8) & (32, 8, 8, 8) & - & - & - \\
    relu1 & gnorm1 & ReLU & (32, 8, 8, 8) & (32, 8, 8, 8) & - & - & - \\
    pool2 & relu1 & MaxPooling & (32, 8, 8, 8) & (32, 4, 4, 4) & (2, 2, 2) & (2, 2, 2) & (0, 0, 0) \\
    conv2 & pool2 & Conv3D & (32, 4, 4, 4) & (64, 2, 2, 2) & (5, 5, 5) & (2, 2, 2) & (2, 2, 2) \\
    gnorm2 & conv2 & GroupNorm & (64, 2, 2, 2) & (64, 2, 2, 2) & - & - & - \\
    relu2 & gnorm2 & ReLU & (64, 2, 2, 2) & (64, 2, 2, 2) & - & - & - \\
    pool3 & relu2 & MaxPooling & (64, 2, 2, 2) & (64, 1, 1, 1) & (2, 2, 2) & (2, 2, 2) & (0, 0, 0) \\
    conv3 & pool3 & Conv3D & (64, 1, 1, 1) & (128, 1, 1, 1) & (1, 1, 1) & (1, 1, 1) & (0, 0, 0) \\
    gnorm3 & conv3 & GroupNorm & (128, 1, 1, 1) & (128, 1, 1, 1) & - & - & - \\
    relu3 & gnorm3 & ReLU & (128, 1, 1, 1) & (128, 1, 1, 1) & - & - & - \\
    flat0 & node feature & Flatten & (128, 1, 1, 1) & (128) & - & - & - \\
\bottomrule
\end{tabular}
}
\vspace{-0.2cm}
\caption{Layer specification for detected object encoder.}
\label{tab:arch_encoder}

\centering
\resizebox{0.8\textwidth}{!}{
\begin{tabular}{c|c|c|c|c}
\toprule
    Child decoder & Input Layer & Type & Input Size & Output Size \\
\midrule
    lin0 & node feature & Linear & 128 & 1280 \\
    relu0 & lin0 & ReLU & 1280 & 1280 \\
    reshape0 & relu0 & Reshape & 1280 & (10, 128) \\
    node\_exist & reshape0 & Linear & (10, 128) & (10, 1) \\
\midrule
    concat0 & (reshape0, reshape0) & Concat. & (10, 128), (10, 128) & (10, 10, 256) \\
    lin1 & concat0 & Linear & (10, 10, 256) & (10, 10, 128) \\
    relu1 & lin1 & ReLU & (10, 10, 128) & (10, 10, 128) \\
    edge\_exist & relu1 & Linear & (10, 10, 128) & (10, 10, 1) \\
\midrule
    mp & (relu1, edge\_exist, reshape0) & Mes. Passing & (10, 10, 128), (10, 10, 1), (10, 128) & (10, 384) \\
    lin2 & mp & Linear & (10, 384) & (10, 128) \\
    relu2 & lin2 & ReLU & (10, 128) & (10, 128) \\
    node\_sem & relu2 & Linear & (10, 128) & (10, \#classes) \\
\midrule
    lin3 & relu2 & Linear & (10, 128) & (10, 128) \\
    relu3 & lin3 & ReLU & (10, 128) & (10, 128) \\
\bottomrule
\end{tabular}
}
\vspace{-0.2cm}
\caption{Layer specification for decoding an object into a part tree.}
\label{tab:arch_child_decoder}

\centering
\resizebox{0.8\textwidth}{!}{
\begin{tabular}{c|c|c|c|c|c|c|c}
\toprule
    Prior refiner & Input Layer & Type & Input Size & Output Size & Kernel Size & Stride & Padding \\
\midrule
    concat0 & (prior, scan occ. grid) & Concat. & (1, 32, 32, 32), (1, 32, 32, 32) & (2, 32, 32, 32) & - & - & - \\
    conv0 & concat0 & Conv3D & (2, 32, 32, 32) & (8, 32, 32, 32) & (3, 3, 3) & (1, 1, 1) & (1, 1, 1) \\
    bnorm0 & conv0 & BatchNorm & (8, 32, 32, 32) & (8, 32, 32, 32) & - & - & - \\
    relu0 & bnorm0 & ReLU & (8, 32, 32, 32) & (8, 32, 32, 32) & - & - & - \\
    conv1 & relu0 & Conv3D & (8, 32, 32, 32) & (16, 32, 32, 32) & (3, 3, 3) & (1, 1, 1) & (1, 1, 1) \\
    bnorm1 & conv1 & BatchNorm & (16, 32, 32, 32) & (16, 32, 32, 32) & - & - & - \\
    relu1 & bnorm1 & ReLU & (16, 32, 32, 32) & (16, 32, 32, 32) & - & - & - \\
    conv2 & relu1 & Conv3D & (16, 32, 32, 32) & (8, 32, 32, 32) & (3, 3, 3) & (1, 1, 1) & (1, 1, 1) \\
    bnorm2 & conv2 & BatchNorm & (8, 32, 32, 32) & (8, 32, 32, 32) & - & - & - \\
    relu2 & bnorm2 & ReLU & (8, 32, 32, 32) & (8, 32, 32, 32) & - & - & - \\
    conv3 & relu2 & Conv3D & (8, 32, 32, 32) & (1, 32, 32, 32) & (1, 1, 1) & (1, 1, 1) & (0, 0, 0) \\
    add3 & (prior, conv3) & Add & (1, 32, 32, 32), (1, 32, 32, 32) & (1, 32, 32, 32) & - & - & - \\
    sigmoid3 & add3 & Sigmoid & (1, 32, 32, 32) & (1, 32, 32, 32) & - & - & - \\
\bottomrule
\end{tabular}
}
\vspace{-0.2cm}
\caption{Layer specification for final part mask refinement.}
\label{tab:arch_decoder}
\end{table*}

\section{Network Architecture Details}
\label{sec:app_network}

We detail our network architecture specification in Tables~\ref{tab:arch_encoder}-\ref{tab:arch_child_decoder}.
Table~\ref{tab:arch_encoder} describes the layers for encoding the detected objects to a feature code. 
The feature code is then input to a decoder which predicts the part tree, as detailed in Table~\ref{tab:arch_decoder}; here, the output of the last layer, \texttt{lin3}, represents a tuple of children latent codes, which predict part prior weights, as specified in Section 3.4 of the main paper.
The final part refinement is then described in Table~\ref{tab:arch_child_decoder}.
Our volumetric object encoder and part refinement are fully convolutional, while the part tree prediction operates on the latent feature representations of shapes and parts with MLP structure.

\section{Additional Baseline Training Details}
\label{sec:app_baselines}

In all our experiments in comparison with state of the art, we leveraged a combination of various approaches.
For the task of Semantic Part Completion, we performed scan completion with SG-NN~\cite{dai2020sgnn} and object detection with MLCVNet~\cite{xie2020mlcvnet}. 
Our UNet baseline is developed as a baseline without any part tree or geometric part prior inference; it consists of only a 3D voxel encoder (four convolutional blocks consisting of 3D convolution, Group Normalization, ReLU activation) and 3D voxel decoder (five convolutional blocks consisting of 3D transposed convolution, 3D convolution, Group Normalization, ReLU activation) with 45 output feature channels, corresponding to binary masks for each part type, and trained with a binary cross entropy loss. 
Without the explicit part structure representations, this UNet baseline tends to predict noisy part masks, or part types from incorrect classes which remain functionally different. 

Note that for experiments with StructureNet~\cite{mo2019structurenet}, we used the same experimental setup as described in their original paper, training different models for each class category.
Since StructureNet operates in the canonical space of the objects, we provided our predicted object orientations from our approach to guide the StructureNet predictions.

\section{Part Types}
\label{sec:app_parts}

\begin{figure*}
\begin{center}
	\includegraphics[width=0.9\linewidth]{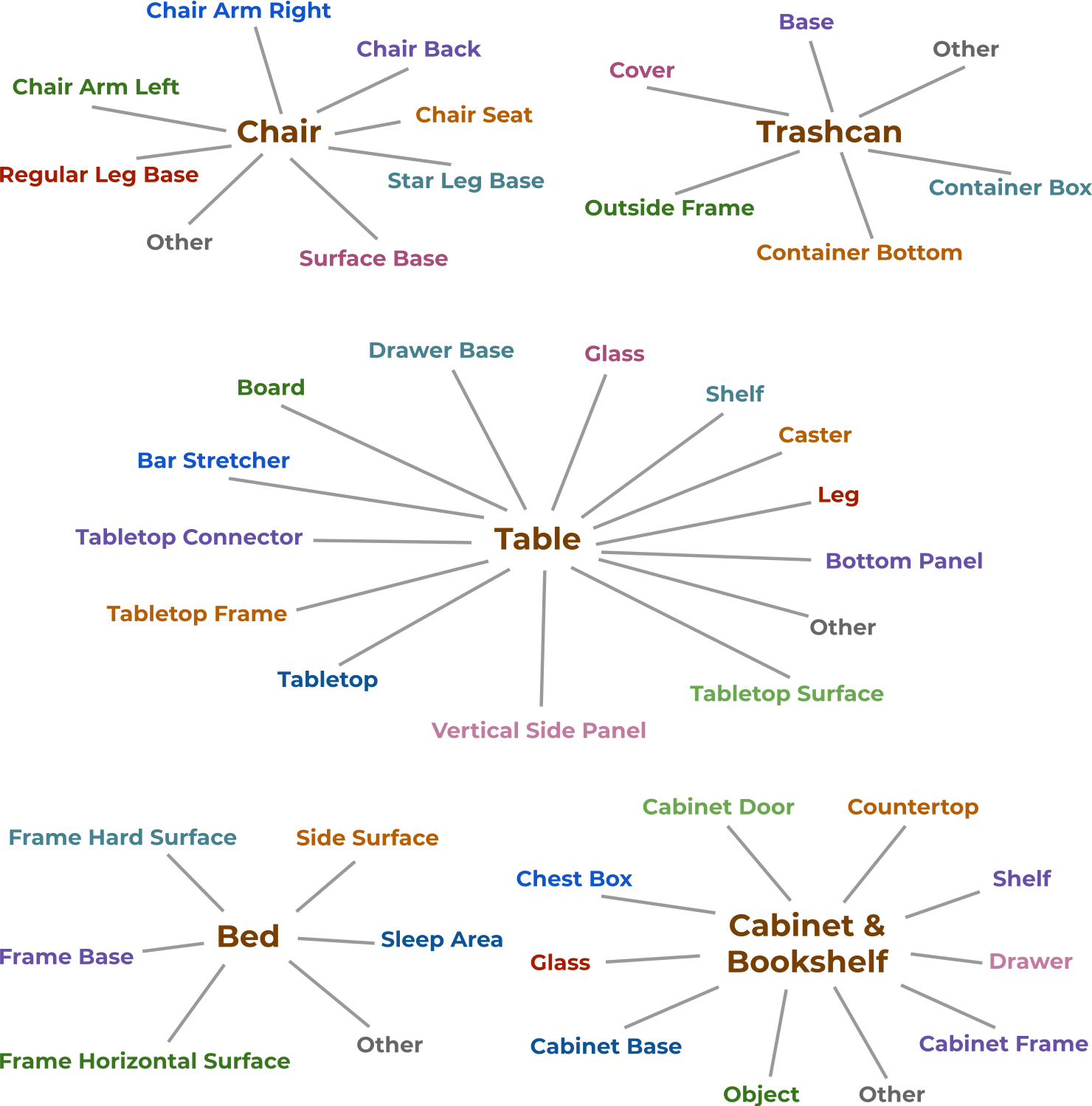}
	\vspace{-0.2cm}
    \caption{Part specification for the parts used in our approach. Note that `cabinet' and `bookshelf' classes have the same set of parts.}
\label{fig:part_spec}
\end{center}
\end{figure*}

In Figure~\ref{fig:part_spec}, we visualize all part types which we trained on. Note that the classes 'cabinet' and 'bookshelf' share the same set of parts, so we use the same part types and priors.

\section{Additional Results}
\label{sec:app_results}

\paragraph{Additional Quantitative Results}

In Table~\ref{tab:inst_cmpl_map_comparison} we additionally evaluate object instance completion using an mAP@25 metric, in comparison to state-of-the-art RevealNet~\cite{hou2020revealnet} and a combination of MLCVNet~\cite{xie2020mlcvnet} with StructureNet~\cite{mo2019structurenet}.
Additionally, in Table~\ref{tab:part_cmpl_comparison_gt}, we evaluate our approach with ground truth 3D detection, i.e., ground truth oriented 3D bounding boxes for each object in the scene. 
Under ground truth detection, our structural part priors enable more robust part decomposition than StructureNet~\cite{mo2019structurenet}.

\paragraph{Additional Part Prior Visualizations}

\begin{figure*}
\begin{center}
	\includegraphics[width=0.8\linewidth]{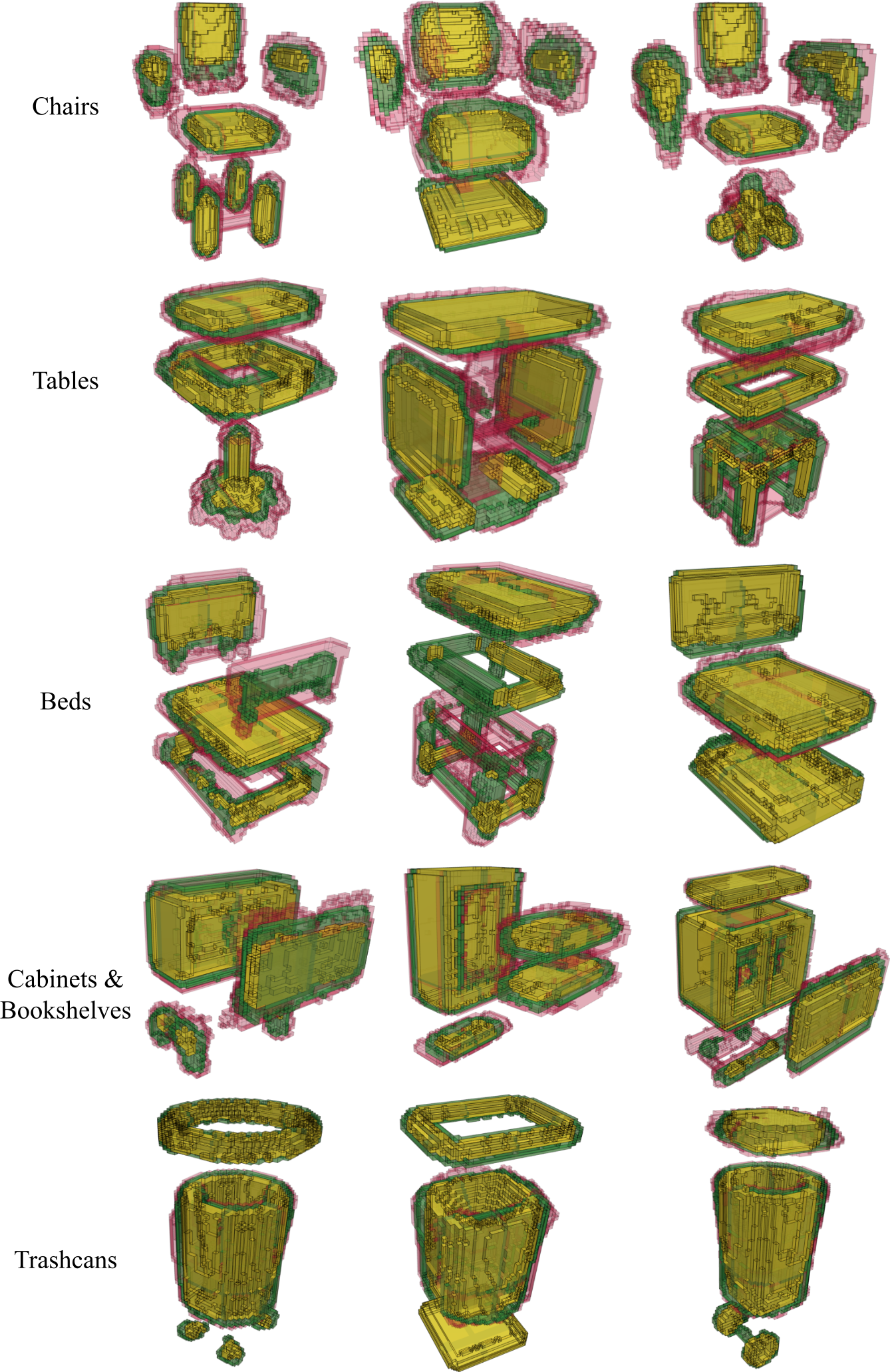}
	\vspace{-0.2cm}
    \caption{Visualization of various part priors.}
\label{fig:more_priors}
\end{center}
\end{figure*}

We show additional examples of computed part priors for each object class category in  Figure~\ref{fig:more_priors}. All  priors are visualized with three level-sets.

\paragraph{Additional Qualitative Semantic Part Completion Results}

\begin{figure*}
\begin{center}
	\includegraphics[width=0.8\linewidth]{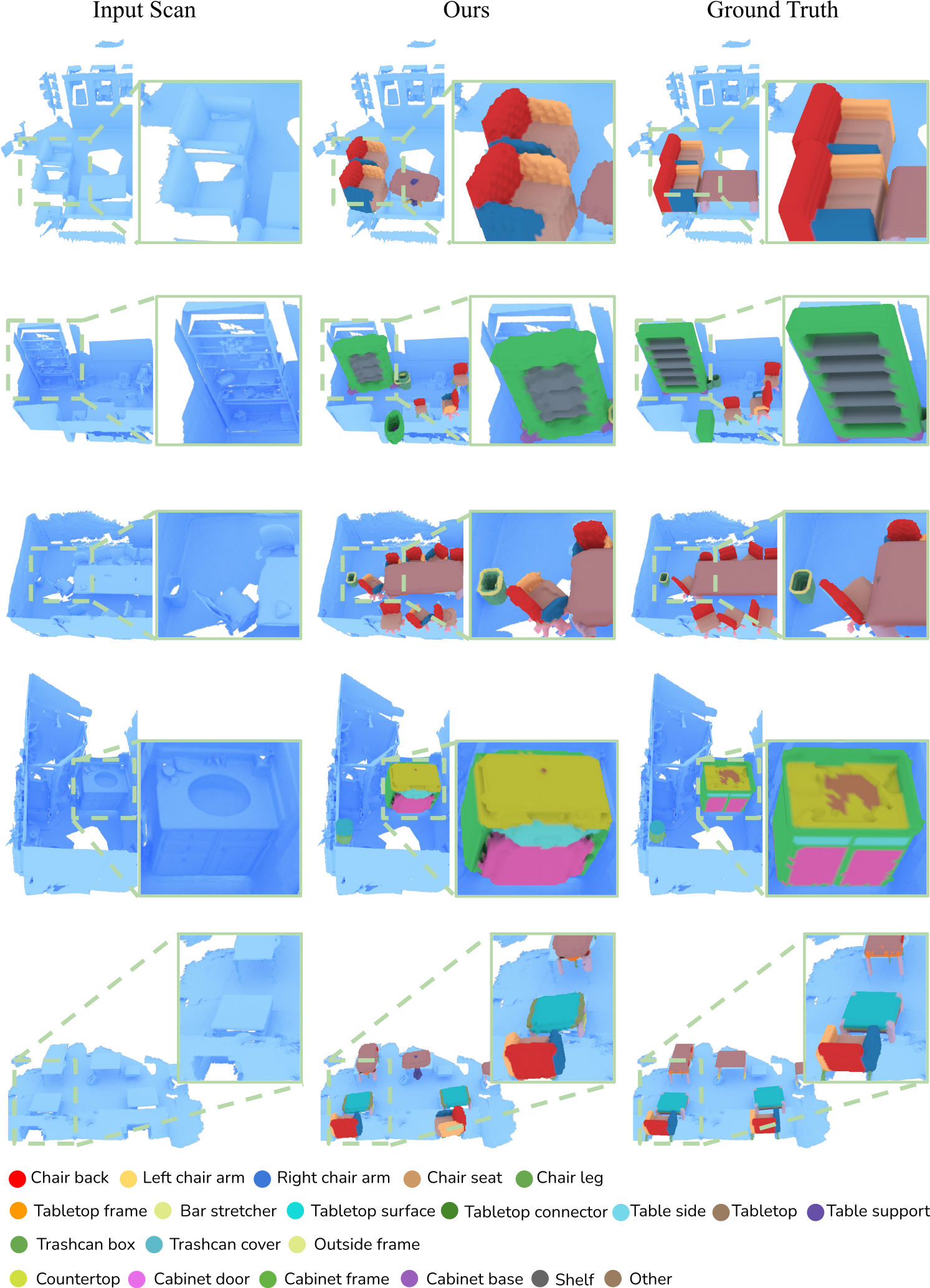}
	\vspace{-0.2cm}
    \caption{Additional qualitative results for our method on ScanNet~\cite{dai2017scannet} scenes and ground truth from Scan2CAD~\cite{avetisyan2019scan2cad} and PartNet~\cite{mo2019partnet}.}
\label{fig:extra_gallery}
\end{center}
\end{figure*}

Figure~\ref{fig:extra_gallery} shows additional examples of our predictions compared with ground-truth. Our method predicts  meaningful part completion across a variety of object categories.

\end{document}